\documentclass{IEEEtran}
\usepackage{graphicx} 
\usepackage{hyperref}
\usepackage{url}
\usepackage[square,sort&compress,numbers]{natbib}
\usepackage{amsmath}
\usepackage{amssymb}
\usepackage{array}
\usepackage{color}
\usepackage{booktabs}
\usepackage{array, caption, threeparttable}
\usepackage[font=small,labelfont=bf,labelsep=none]{caption}

\usepackage{stfloats}
\usepackage{subfigure}
\usepackage{cuted}
\title{Edge-Boundary-Texture Loss: A Tri-Class Generalization of Weighted Binary Cross-Entropy for Enhanced Edge Detection}
\author{Hao Shu\thanks{Hao Shu: Hao$\_$B$\_$Shu@163.com}\IEEEauthorrefmark{1}
\\
\IEEEauthorblockA{\IEEEauthorrefmark{1} Sun-Yat-Sen University, Shenzhen, China.}
\\
\IEEEauthorblockA{\IEEEauthorrefmark{1} Shenzhen University, Shenzhen, China.}
}
\date{}

\begin{document}

\maketitle

\begin{abstract}

Edge detection (ED) remains a fundamental task in computer vision, yet its performance is often hindered by the ambiguous nature of non-edge pixels near object boundaries. The widely adopted Weighted Binary Cross-Entropy (WBCE) loss treats all non-edge pixels uniformly, overlooking the structural nuances around edges and often resulting in blurred predictions. In this paper, we propose the Edge-Boundary-Texture (EBT) loss, a novel objective that explicitly divides pixels into three categories, edge, boundary, and texture, and assigns each a distinct supervisory weight. This tri-class formulation enables more structured learning by guiding the model to focus on both edge precision and contextual boundary localization. We theoretically show that the EBT loss generalizes the WBCE loss, with the latter becoming a limit case. Extensive experiments across multiple benchmarks demonstrate the superiority of the EBT loss both quantitatively and perceptually. Furthermore, the consistent use of unified hyperparameters across all models and datasets, along with robustness to their moderate variations, indicates that the EBT loss requires minimal fine-tuning and is easily deployable in practice.

\end{abstract}

\section{Introduction}

Edge detection (ED) remains a cornerstone problem in computer vision, providing essential structure-aware cues for tasks such as segmentation~\cite{NN2019EdgeConnect}, recognition~\cite{ZD2007An}, and scene understanding~\cite{MR2011Edge}. Modern deep learning-based ED models have achieved impressive results by leveraging multi-scale features and supervised learning, with representative works such as HED~\cite{XT2015Holistically}, BDCN~\cite{HZ2022BDCN}, Dexi~\cite{SS2023Dense}, and the E-S paradigm~\cite{S2025Boost}. However, despite these architectural advances, the quality of edge predictions—especially in terms of sharpness and spatial precision—remains limited by the shortcomings of commonly used loss functions.

Most ED models adopt the Weighted Binary Cross-Entropy (WBCE) loss to counteract the severe imbalance between edge and non-edge pixels. While WBCE improves recall by emphasizing edge regions during training, it fails to distinguish structurally important pixels near the edges from distant background textures. As a result, predicted edge maps often appear thick and blurry, with uncertain boundaries that require additional post-processing such as non-maximum suppression (NMS) to achieve visually acceptable results. This mismatch between the training objective and the desired perceptual output undermines the clarity and reliability of ED results.

Although several methods have been proposed to mitigate this issue~\cite{DS2018Learning,CK2024RankED}, they often involve trade-offs between quantitative scores and perceptual quality, and can be difficult to deploy due to complex hyperparameter tuning across different datasets and models. Therefore, designing an effective and easy-to-use loss function that produces crisp and clear edge predictions remains a significant challenge.

To address this, we propose a novel loss function called the Edge-Boundary-Texture (EBT) loss. Instead of treating ED as a binary classification task, the EBT loss introduces a three-class supervision scheme that explicitly separates image pixels into edge, boundary, and texture regions. This finer-grained categorization allows the model not only to prioritize true edges but also to pay structured attention to nearby boundary areas, which are often perceptually important yet poorly captured by traditional losses. The EBT loss assigns adaptive weights to each region based on class balance and spatial relevance, guiding the model toward crisper and more spatially localized edge predictions.

Moreover, we show that the EBT loss is a strict generalization of the WBCE loss: when the boundary region expands to cover the entire image, the EBT loss asymptotically reduces to WBCE. This theoretical property establishes a principled connection between the traditional WBCE and the proposed EBT loss, positioning EBT as a more structured alternative rather than an ad-hoc replacement.

Extensive experiments on multiple ED benchmarks demonstrate that integrating the EBT loss into existing ED models improves both visual quality and evaluation metrics, especially under strict evaluation protocols, where perceptual sharpness and spatial precision are critical. In summary, the contributions are as follows:

\begin{itemize}
\item We propose the EBT loss, a novel loss function that introduces a structured supervision strategy based on edge, boundary, and texture pixel classes.
\item We provide a formal definition of the EBT loss and show that WBCE is a special case of it by asymptotic analysis, establishing its theoretical generality.
\item We empirically validate the effectiveness of the EBT loss across multiple models and datasets, demonstrating consistent improvements in both prediction quality and quantitative performance.
\end{itemize}

\section{Related Works}

This section surveys the landscape of ED research, spanning datasets, model designs, loss functions, and evaluation methodologies. While the field has made notable progress, persistent challenges continue to drive innovation.

\subsection{Datasets}

Early advancements in ED heavily leaned on datasets developed for broader vision tasks, such as segmentation and contour extraction. Datasets like BSDS300 and BSDS500~\cite{MF2001A}, originally designed for segmentation tasks with multiple human annotations, have remained staples in ED benchmarking. Additional sources, including MDBD~\cite{MK2016A}, NYUD~\cite{SH2012Indoor}, and PASCAL VOC~\cite{EV2010The}, have also been adapted to support ED tasks, despite not being purpose-built for this objective.

In response to the limitations of these general-purpose datasets, the community introduced datasets specifically tailored for ED. For example, BIPED and BIPED2~\cite{SR2020Dense} provided high-resolution, single-instance edge labels with improved clarity. BRIND~\cite{PH2021RINDNet} went further by labeling edge types, such as reflectance and depth, enhancing both richness and complexity. More recently, UDED~\cite{SL2023Tiny} emphasized precision in labeling over dataset size, prioritizing annotation consistency and quality.

Despite these advances, human annotations remain prone to subjectivity and noise. Issues such as label inconsistency and semantic ambiguity continue to impact both model training and evaluation. Recent studies have proposed strategies to mitigate these effects~\cite{FG2023Practical, WD2024One, S2025Enhancing}, but annotation noise remains a fundamental bottleneck for achieving reliable generalization.

\subsection{Modeling Approaches}

The evolution of ED models spans three distinct eras: traditional edge operators, classical machine learning approaches, and deep learning-based architectures.

Traditional edge detectors like the Sobel filter~\cite{K1983On} and the Canny detector~\cite{C1986A} rely on handcrafted gradients and thresholds. While fast and interpretable, their performance breaks down in the presence of texture, shadows, or complex object boundaries.

Later, statistical learning models incorporated engineered features such as shape descriptors~\cite{MF2004Learning}, sketch tokens~\cite{LZ2013Sketch}, and contour histograms~\cite{AM2011Contour}, paired with classifiers like random forests~\cite{DZ2015Fast} and logistic regression~\cite{R2008Multi}. Though more robust than their handcrafted predecessors, these methods lacked scalability and failed to adapt across diverse scenes.

Deep learning brought a transformative shift from traditional methods to data-driven end-to-end learning. Pioneering models like HED~\cite{XT2015Holistically} leveraged deep supervision to combine multi-scale features. Subsequent models, including RCF~\cite{LC2017Richer} and BDCN~\cite{HZ2022BDCN}, refined feature fusion and deep-layer activation schemes. Architectures such as PiDiNet~\cite{SL2021Pixel} and DexiNet~\cite{SS2023Dense} targeted computational efficiency and prediction precision, respectively. More recently, attention-based designs like EDTER~\cite{PH2022Edter} and EdgeNAT~\cite{JG2024EdgeNAT} explored long-range dependency modeling through transformers in edge-related tasks. In parallel, diffusion-based frameworks~\cite{ZH2024Generative, YX2024Diffusionedge} have emerged, though their benefits over CNNs for ED are still under active investigation. Another promising trend is the Extractor-Selector paradigm~\cite{S2025Boost}, which separates feature extraction and refines feature selection, providing new insights for enhancing ED models.

\subsection{Loss Function Design}

Loss functions play a pivotal role in determining how models learn from imbalanced pixel distributions, especially given the sparse nature of edge labels. The standard Binary Cross-Entropy (BCE) loss often biases models toward background predictions, leading to high precision but low recall. To mitigate this, the Weighted BCE (WBCE) loss introduces class-balancing terms that reweight edge and non-edge pixels based on their frequency~\cite{XT2015Holistically}.

Nevertheless, WBCE has limitations. It tends to produce overly thick edge maps, requiring post-processing such as NMS to sharpen predictions. Alternative losses have been developed to address these issues. Dice loss~\cite{DS2018Learning} maximizes overlap between predicted and ground-truth regions, while region-based losses like tracing loss~\cite{HX2022Unmixing} focus on spatial consistency around edges. Ranking-based objectives~\cite{CK2024RankED, CL2019Towards} promote better precision by explicitly modeling the relative ordering of pixel values. Additionally, uncertainty-aware formulations like Symmetrization WBCE (SWBCE) loss~\cite{S2025Rethinking} account for human labeling imbalance based on predicted maps, while binarization-aware designs such as BAA loss function~\cite{S2025Binarization} focus on the mismatch of model training and evaluations.

Although these innovations have improved aspects of perceptual quality and structure, they often introduce trade-offs such as increased training instability or reliance on sensitive hyperparameters. The search continues for a principled loss that balances precision, recall, and edge sharpness in a stable, generalizable, and easily employable manner.

\subsection{Evaluation Protocols}

Evaluating ED performance typically relies on the F$_1$-score, using metrics such as:

\begin{itemize}
\item \textbf{ODS (Optimal Dataset Score):} Measures the best F$_1$ score using the optimal threshold across the entire dataset.
\item \textbf{OIS (Optimal Image Score):} Averages the best F$_1$ scores for each image using individualized optimal thresholds.
\item \textbf{AP (Average Precision):} Evaluates the mean precision-recall performance.
\end{itemize}

These metrics employ a tolerance-based matching strategy to allow spatial deviations between predicted and annotated edges~\cite{MF2004Learning}. However, the definition of tolerance, often expressed as a radius that scales with image resolution, varies across datasets and can significantly affect comparability. More recent studies advocate for standardized, stricter matching protocols~\cite{S2025Enhancing}, such as using a fixed 1-pixel radius, to ensure consistency and reduce error effects. Nonetheless, as a newly presented suggestion, this protocol has only been adopted in the most recent works.

\section{Methodology}

\subsection{Motivation}

Given a predicted edge probability map $\hat{Y}$ and a binary ground-truth edge map $Y$, early ED approaches commonly frame the task as binary classification, applying the Binary Cross-Entropy (BCE) loss to distinguish edge from non-edge pixels:

\begin{equation}
\begin{aligned}
	L_{BCE}(\hat{Y},Y)=&\frac{-\sum_{i\in Y}[y_{i}log(\hat{y_{i}})+(1-y_{i})log(1-\hat{y_{i}})]}{|Y|}\\
\end{aligned}
\end{equation}

Here, $\hat{y}_i$ and $y_i$ represent the predicted and ground-truth labels for pixel $i$, respectively.

While BCE loss achieves high precision, it often suffers from low recall due to the severe imbalance between edge and non-edge pixels. To address this, the Weighted Binary Cross-Entropy (WBCE) loss introduces class weights to better balance the training signal:

\begin{equation}
\begin{aligned}
	&L_{WBCE}(\hat{Y},Y)\\=
    &\frac{-\sum_{i\in Y}\alpha_{i}[y_{i}log(\hat{y_{i}})+(1-y_{i})log(1-\hat{y_{i}})]}{|Y|}\\
    =&\frac{-\sum_{i\in Y^{+}}\alpha_{i}log(\hat{y_{i}})-\sum_{i\in Y^{-}}\alpha_{i}log(1-\hat{y_{i}})}{|Y|}\\
    =&\frac{-\alpha\sum_{i\in Y^{+}}log(\hat{y_{i}})-\lambda(1-\alpha)\sum_{i\in Y^{-}}log(1-\hat{y_{i}})}{|Y|}
\end{aligned}
\end{equation}
where,
\begin{itemize}
	\item $Y^{+}$ and $Y^{-}$ denote the sets of edge and non-edge pixels in $Y$, respectively.
    
	\item The weights $\alpha_i$ scale loss contributions based on pixel class, with $\lambda$ acting as a balancing hyperparameter: For $i \in Y^{+}$, $\alpha_{i} = \alpha = \frac{|Y^{-}|}{|Y|}$, while for $i \in Y^{-}$, $\alpha_{i} = \lambda(1 - \alpha)$, and $\lambda$ is usually set to 1.1 as default.
\end{itemize}

Despite mitigating class imbalance, WBCE still overlooks spatial structure. It treats all non-edge pixels equally, regardless they lie adjacent to edges or deep within homogeneous background regions. In practice, pixels near edges share visual characteristics with true edges and are more likely to be misclassified. However, WBCE assigns them the same low weight as distant background pixels, leading to insufficient supervision. As a result, predictions often appear blurry and lack precise delineation.

To address this, we propose the Edge-Boundary-Texture (EBT) loss, which introduces a three-class pixel taxonomy to provide more structured and effective guidance. Unlike WBCE, which dichotomizes pixels as edge or non-edge, EBT loss further distinguishes boundary and texture pixels in non-edge regions, enhancing prediction sharpness and spatial awareness.

\subsection{Formal Definition of EBT Loss}

To define the EBT loss, we categorize pixels into three semantically meaningful categories:
\begin{itemize}
    \item \textbf{Edge pixels} ($Y_E$): Pixels explicitly labeled as edges in the ground-truth.
    \item \textbf{Boundary pixels} ($Y_B$): Non-edge pixels that lie within a square window of size $(2r+1) \times (2r+1)$ centered at any edge pixel, where $r$ is a pre-fixed boundary radius hyperparameter. $r=7$ is set as the default, chosen by experiments.
    \item \textbf{Textural pixels} ($Y_T$): Remaining pixels that are neither edge nor boundary ones.
\end{itemize}

This categorization enables the model to better differentiate structurally significant edge pixels, semantically adjacent boundary regions prone to confusion with edges, and low-salience textural backgrounds, thereby reducing prediction blur. Formally, the EBT loss is defined as:

\begin{strip}
\begin{equation}
\begin{aligned}
	&L_{EBT}(\hat{Y},Y)\\
    &=\frac{-\sum_{i\in Y}\hat{\alpha}_{i}[y_{i}log(\hat{y_{i}})+(1-y_{i})log(1-\hat{y_{i}})]}{|Y|}=\frac{-\sum_{i\in Y_{E}}\hat{\alpha}_{i}log(\hat{y_{i}})-\sum_{i\in Y_{B}}\hat{\alpha}_{i}log(1-\hat{y_{i}})-\sum_{i\in Y_{T}}\hat{\alpha}_{i}log(1-\hat{y_{i}})}{|Y|}\\
     &=\frac{-\sum_{i\in Y_{E}}B_{E}W_{E}log(\hat{y_{i}})-\sum_{i\in Y_{B}}B_{B}W_{B}log(1-\hat{y_{i}})-\sum_{i\in Y_{T}}B_{T}W_{T}log(1-\hat{y_{i}})}{|Y|}
\end{aligned}
\end{equation}
\end{strip}

\noindent where:
\begin{itemize}
    \item $\hat{\alpha}_i$ is the assigned weight of pixel $i$, defined as $B_E W_E$ for edge pixels, $B_B W_B$ for boundary pixels, and $B_T W_T$ for textural pixels.
    \item $B_E$, $B_B$, and $B_T$ are pre-defined balancing coefficients that refine the relative importance of each class.
    \item $W_E$, $W_B$, and $W_T$ are adaptive weights designed to balance the size of the classes, defined as:
    \begin{equation}
        W_E = \frac{|Y_B \cup Y_T|}{|Y|},
        W_B = \frac{|Y_E \cup Y_T|}{|Y|},
        W_T = \frac{|Y_E \cup Y_B|}{|Y|}
    \end{equation}
These terms compensate for per-image class imbalance and ensure that smaller pixel groups (e.g., edges) receive proportionally stronger supervision.
\end{itemize}

Compared to WBCE, EBT offers finer supervision by explicitly modeling the challenging boundary zone. This guides the network to better preserve the surrounding structure while localizing edges, resulting in sharper, more accurate predictions.

\subsection{Asymptotic Behavior of EBT Loss}

The EBT loss can be viewed as a generalization of the WBCE loss from a mathematical perspective. When $r \rightarrow +\infty$, or more practically, when $r \geq \max\{R_H, R_W\}-1$, where $R_H$ and $R_W$ are the maximum image height and width, the three classes collapse as follows:
\begin{equation}
    Y_{E}=Y^{+},\qquad
    Y_{B}=Y^{-},\qquad
    Y_{T}=\emptyset
\end{equation}
Furthermore, the corresponding adaptive weights simplify to:
\begin{equation}
    W_{E}=\frac{|Y^{-}|}{|Y|}=\alpha,
    W_{B}=\frac{|Y^{+}|}{|Y|}=1-\alpha,
    W_{T}=\frac{|Y|}{|Y|}=1
\end{equation}
Therefore, by additionally setting $B_E=1$, $B_B=\lambda$, the EBT loss reduces to:

\begin{strip}
\begin{equation}
\begin{aligned}
  L_{EBT}(\hat{Y},Y)&=\frac{-\sum_{i\in Y_{E}}B_{E}W_{E}log(\hat{y_{i}})-\sum_{i\in Y_{B}}B_{B}W_{B}log(1-\hat{y_{i}})-\sum_{i\in Y_{T}}B_{T}W_{T}log(1-\hat{y_{i}})}{|Y|}   \\
  &=\frac{-\sum_{i\in Y^{+}}\alpha log(\hat{y_{i}})-\sum_{i\in Y^{-}}\lambda(1-\alpha)log(1-\hat{y_{i}})}{|Y|}=L_{WBCE}(\hat{Y},Y)\qquad (r\rightarrow +\infty)
\end{aligned}
\end{equation}
\end{strip}

This demonstrates that WBCE is a special case of EBT loss under extreme spatial inclusion, affirming EBT as a principled generalization. It introduces structured spatial supervision while remaining mathematically consistent with WBCE.

\section{Experiment}

We systematically evaluate the performance of the proposed EBT loss by benchmarking it against the widely used WBCE loss across multiple ED models and datasets. An ablation study is also conducted to demonstrate the stability of the EBT hyperparameters.

\subsection{Experimental Settings}

\subsubsection{Evaluation Scope and Model Setup}

The experiments cover four ED models: HED~\cite{XT2015Holistically}, BDCN~\cite{HZ2022BDCN}, Dexi~\cite{SS2023Dense}, and EES3~\cite{S2025Boost}, where EES3 represents the enhanced E-S architecture that employs the other three models as parallel extractors within the standard selector framework.

Experiments are conducted on five diverse datasets: BIPED2~\cite{SR2020Dense}, BRIND~\cite{PH2021RINDNet}, BSDS500~\cite{MF2001A}, UDED~\cite{SL2023Tiny}, and NYUD2~\cite{SH2012Indoor}\footnote{Here, BIPED2, BRIND, and UDED are datasets specifically for ED, while BSDS500 and NYUD2 are ordinarily for segmentation tansks. Therefore, the EBT loss is in fact tested on both ED tasks and segmentation tasks.}. Each model is retrained independently on each dataset to ensure fair comparisons. For EES3, only the selector is retrained, while the extractors use pretrained weights from WBCE-trained models as provided in~\cite{S2025Boost}.

To isolate the effect of the loss function, we adopt a direct substitution strategy: the WBCE loss in each baseline model is replaced with the proposed EBT loss, while keeping all network weights and architectural configurations unchanged, in the experiments with EBT loss. Here, no additional tuning is performed to optimize the EBT layer weights for individual architectures. This might result in the sub-optimality of model performances with EBT loss, but it is sufficient to demonstrate the superiority of the EBT loss.

Unless otherwise specified, the balancing hyperparameters for edge ($B_E$), boundary ($B_B$), and texture ($B_T$) pixel weights are fixed at 1.0, 0.8, and 0.5, respectively.

Evaluation is based on the standard ODS, OIS, and AP metrics~\cite{MF2004Learning} under the strictest 1-pixel error tolerance without post-processing techniques such as NMS to precisely highlight the intrinsic capabilities of the models and loss functions, in line with recent practices~\cite{S2025Enhancing}. Results using relaxed error margins (4.3–11.1 pixels) and NMS are included in the appendix.

\subsubsection{Dataset Preparation and Augmentation Strategy}

Dataset preprocessing and partitioning follow the protocols in~\cite{S2025Boost}, with the following specific splits:

\begin{itemize}
    \item \textbf{BRIND and BSDS500:} 400 images for training, 100 for testing.
    \item \textbf{UDED:} 20 images for training, 7 for testing (2 low-resolution samples are excluded).
    \item \textbf{BIPED2:} 200 training images, 50 test images.
    \item \textbf{NYUD2:} 1100 training images, 300 test images.
\end{itemize}

Image augmentation techniques include:
\begin{itemize}
    \item Progressive resizing: input images are recursively halved until their dimensions fall below $640\times640$ (not applied to NYUD2).
    \item Geometric transformations: each image is rotated at $0^\circ$, $90^\circ$, $180^\circ$, and $270^\circ$, with horizontal flipping applied at each rotation.
    \item Integration of noiseless samples into training, as suggested in \cite{S2025Enhancing}.
\end{itemize}

\subsubsection{Training Protocol}

During training, images are randomly cropped to $320\times320$ and resampled every five epochs. 

All models are trained using the Adam optimizer with a learning rate of $10^{-4}$ and a weight decay of $10^{-8}$, on an A100 GPU. The batch size is set to 8.

Training is performed for 200 epochs on UDED and 50 epochs on all other datasets.

\subsubsection{Inference Procedure}

During inference, each test image is divided into patches of $320\times320$ with a stride of 304 pixels (i.e., 16-pixel overlap). Patch-level predictions are made independently and merged to reconstruct the full edge map.

\subsection{Experimental results}

This subsection provides the main experimental results on the five datasets: BIPED2, BRIND, BSDS500, UDED, and NYUD2, based on the four representative ED models: HED, BDCN, Dexi, and EES3. For each model, we compare results using the standard WBCE loss (from official implementations) and the proposed EBT loss under the strictest 1-pixel error tolerance without NMS, following recent benchmarks. Evaluations under more relaxed benchmarks are provided in the appendix. Metrics include ODS, OIS, and AP.

\begin{table}[htbp]
\renewcommand\arraystretch{1.3}
\centering
\caption{\qquad \textbf{Results on BIPED2 with 1-pixel error tolerance without NMS:} \textit{-EBT} represents that the model is trained with the EBT loss, while without \textit{-EBT} represents that the model is trained with the WBCE loss. \textit{Avg} and \textit{Avg-EBT} represent that the scores are averaged over the four models with WBCE and EBT losses, respectively. The improvements of the EBT loss compared to the WBCE loss are marked in parentheses, and the better scores are \textbf{in bold}.}
\label{Raw-BIPED2}
\begin{tabular}{|p{16.25mm}<{\centering}|p{18.25mm}<{\centering}|p{18.25mm}<{\centering}|p{19.75mm}<{\centering}|}
\hline
    & ODS   & OIS   & AP \\
\hline
HED (2015) & 0.592 & 0.602 & 0.347\\
\hline
HED-EBT & \textbf{0.618 (+4.39$\%$)} & \textbf{0.622 (+3.32$\%$)} & \textbf{0.434 (+25.07$\%$)} \\
\hline
BDCN (2022) & 0.629 & 0.635 & 0.421\\
\hline
BDCN-EBT & \textbf{0.638 (+1.43$\%$)} & \textbf{0.642 (+1.10$\%$)} & \textbf{0.469 (+11.40$\%$)} \\
\hline
Dexi (2023) & 0.632 & 0.636 & 0.432 \\
\hline
Dexi-EBT & \textbf{0.641 (+1.42$\%$)} & \textbf{0.645 (+1.42$\%$)} & \textbf{0.492 (+13.89$\%$)} \\
\hline
EES3 (2025) & 0.659 & 0.663 & 0.473 \\
\hline
EES3-EBT & \textbf{0.675 (+2.43$\%$)} & \textbf{0.679 (+2.41$\%$)} & \textbf{0.541 (+14.38$\%$)} \\
\hline
Avg & 0.628 & 0.634 & 0.418 \\
\hline
Avg-EBT & \textbf{0.643 (+2.39$\%$)} & \textbf{0.647 (+2.05$\%$)} & \textbf{0.484 (+15.79$\%$)} \\
\hline
\end{tabular}
\end{table}

\begin{table}[htbp]
\renewcommand\arraystretch{1.3}
\centering
\caption{\qquad \textbf{Results on BRIND with 1-pixel error tolerance without NMS:} The notations follow Table \ref{Raw-BIPED2}.}
\label{Raw-BRIND}
\begin{tabular}{|p{16.25mm}<{\centering}|p{18.25mm}<{\centering}|p{18.25mm}<{\centering}|p{19.75mm}<{\centering}|}
\hline
    & ODS   & OIS   & AP \\
\hline
HED (2015) & 0.645 & 0.656 & 0.417\\
\hline
HED-EBT & \textbf{0.653 (+1.24$\%$)} & \textbf{0.663 (+1.07$\%$)} & \textbf{0.541 (+29.74$\%$)} \\
\hline
BDCN (2022) & 0.659 & 0.672 & 0.467\\
\hline
BDCN-EBT & \textbf{0.677 (+2.73$\%$)} & \textbf{0.686 (+2.08$\%$)} & \textbf{0.596 (+27.62$\%$)} \\
\hline
Dexi (2023) & 0.666 & 0.672 & 0.478 \\
\hline
Dexi-EBT & \textbf{0.672 (+0.90$\%$)} & \textbf{0.679 (+1.04$\%$)} & \textbf{0.575 (+20.29$\%$)} \\
\hline
EES3 (2025) & 0.679 & 0.688 & 0.576 \\
\hline
EES3-EBT & \textbf{0.683 (+0.59$\%$)} & \textbf{0.691 (+0.44$\%$)} & \textbf{0.600 (+4.17$\%$)} \\
\hline
Avg & 0.662 & 0.672 & 0.485 \\
\hline
Avg-EBT & \textbf{0.671 (+1.36$\%$)} & \textbf{0.680 (+1.19$\%$)} & \textbf{0.578 (+19.18$\%$)} \\
\hline
\end{tabular}
\end{table}

\begin{table}[htbp]
\renewcommand\arraystretch{1.3}
\centering
\caption{\qquad \textbf{Results on BSDS500 with 1-pixel error tolerance without NMS:} The notations follow Table \ref{Raw-BIPED2}. Only label 1 of the annotations is used as the ground-truth.}
\label{Raw-BSDS}
\begin{tabular}{|p{16.25mm}<{\centering}|p{18.25mm}<{\centering}|p{18.25mm}<{\centering}|p{19.75mm}<{\centering}|}
\hline
    & ODS   & OIS   & AP \\
\hline
HED (2015) & 0.420 & 0.433 & 0.175 \\
\hline
HED-EBT & \textbf{0.437 (+4.05$\%$)} & \textbf{0.450 (+3.93$\%$)} & \textbf{0.262 (+49.71$\%$)} \\
\hline
BDCN (2022) & 0.454 & 0.465 & 0.240 \\
\hline
BDCN-EBT & \textbf{0.474 (+4.41$\%$)} & \textbf{0.486 (+4.52$\%$)} & \textbf{0.298 (+24.17$\%$)}  \\
\hline
Dexi (2023) & 0.465 & 0.473 & 0.242 \\
\hline
Dexi-EBT & \textbf{0.475 (+2.15$\%$)} & \textbf{0.490 (+3.59$\%$)} & \textbf{0.303 (+25.21$\%$)} \\
\hline
EES3 (2025) & 0.486 & 0.500 & 0.246 \\
\hline
EES3-EBT & \textbf{0.492 (+1.23$\%$)}& \textbf{0.507 (+1.40$\%$)}& \textbf{0.331 (+34.55$\%$)} \\
\hline
Avg & 0.456 & 0.468 & 0.226 \\
\hline
Avg-EBT & \textbf{0.470 (+3.07$\%$)} & \textbf{0.482 (+2.99$\%$)} & \textbf{0.299 (+32.30$\%$)} \\
\hline
\end{tabular}
\end{table}

\begin{table}[htbp]
\renewcommand\arraystretch{1.3}
\centering
\caption{\qquad \textbf{Results on UDED with 1-pixel error tolerance without NMS:} The notations follow Table \ref{Raw-BIPED2}. The 1-pixel error tolerance is with regard to the test image with the lowest resolution.}
\label{Raw-UDED}
\begin{tabular}{|p{16.25mm}<{\centering}|p{18.25mm}<{\centering}|p{18.25mm}<{\centering}|p{19.75mm}<{\centering}|}
\hline
    & ODS   & OIS   & AP \\
\hline
HED (2015) & \textbf{0.716} & \textbf{0.751} & 0.489 \\
\hline
HED-EBT & 0.713 (-0.42$\%$) & 0.739 (-1.60$\%$) & \textbf{0.502 (+2.66$\%$)} \\
\hline
BDCN (2022) & 0.706 & 0.744 & 0.530 \\
\hline
BDCN-EBT & \textbf{0.733 (+3.82$\%$)}& \textbf{0.766 (+2.96$\%$)} & \textbf{0.571 (+7.74$\%$)} \\
\hline
Dexi (2023) & \textbf{0.733} & \textbf{0.752} & 0.600 \\
\hline
Dexi-EBT & 0.724 (-1.23$\%$) & 0.746 (-0.80$\%$) & \textbf{0.607 (+1.17$\%$)} \\
\hline
EES3 (2025) & 0.751 & 0.778 & 0.601 \\
\hline
EES3-EBT & \textbf{0.775 (+3.20$\%$)}& \textbf{0.802 (+3.08$\%$)} & \textbf{0.715 (+18.97$\%$)} \\
\hline
Avg & 0.727 & 0.756 & 0.555 \\
\hline
Avg-EBT & \textbf{0.736 (+1.24$\%$)} & \textbf{0.763 (+0.93$\%$)} & \textbf{0.599 (+7.93$\%$)} \\
\hline
\end{tabular}
\end{table}

\begin{table}[htbp]
\renewcommand\arraystretch{1.3}
\centering
\caption{\qquad \textbf{Results on NYUD2 with 1-pixel error tolerance without NMS:} The notations follow Table \ref{Raw-BIPED2}. The Canny algorithm is run to extract the edge labels from the segmentation annotations.}
\label{Raw-NYUD2}
\begin{tabular}{|p{16.25mm}<{\centering}|p{18.25mm}<{\centering}|p{18.25mm}<{\centering}|p{19.75mm}<{\centering}|}
\hline
    & ODS   & OIS   & AP \\
\hline
HED (2015) & 0.376 & 0.386 & 0.146 \\
\hline
HED-EBT & \textbf{0.377 (+0.27$\%$)}& \textbf{0.393 (+1.81$\%$)} & \textbf{0.166 (+13.70$\%$)} \\
\hline
BDCN (2022) & \textbf{0.406} & \textbf{0.414} & 0.183 \\
\hline
BDCN-EBT & 0.402 (-0.99$\%$) & 0.413 (-0.24$\%$) & \textbf{0.199 (+8.74$\%$)} \\
\hline
Dexi (2023) & \textbf{0.408} & \textbf{0.416} & 0.161 \\
\hline
Dexi-EBT & 0.405 (-0.74$\%$) & 0.413 (-0.72$\%$) & \textbf{0.207 (+28.57$\%$)} \\
\hline
EES3 (2025) & 0.408 & 0.418 & 0.157 \\
\hline
EES3-EBT & \textbf{0.412 (+0.98$\%$)} & \textbf{0.422 (+0.96$\%$)} & \textbf{0.199 (+26.75$\%$)} \\
\hline
Avg & \textbf{0.401} & 0.409 & 0.162 \\
\hline
Avg-EBT & 0.399 (-0.50$\%$) & \textbf{0.410 (+0.24$\%$)} & \textbf{0.193 (+19.14$\%$)} \\
\hline
\end{tabular}
\end{table}

\subsubsection{Quantitative Results}

The quantitative results are summarized in Tables \ref{Raw-BIPED2} to \ref{Raw-NYUD2}. Across all datasets, the EBT loss delivers comparable or superior performance, especially in terms of AP, which reflects the model's ability to make fine-grained distinctions between edge and non-edge regions.

On BIPED2 (Table \ref{Raw-BIPED2}), the EBT loss significantly boosts AP across all four models. The average AP increases from 0.418 to 0.484, marking a 15.79$\%$ improvement. ODS and OIS also show consistent gains, with average improvements of 2.39\% and 2.05\%, respectively.

On BRIND (Table \ref{Raw-BRIND}), the EBT-enhanced models achieve consistent improvements in all metrics, with the average AP improving 19.18\%. HED and BDCN show especially large AP gains (up to 29.74\% and 27.62\%, respectively). Also, ODS and OIS metrics are improved by EBT loss consistently.

The trend continues on BSDS500 (Table \ref{Raw-BSDS}), where EBT provides a substantial 32.30\% improvement in average AP. This is particularly notable considering the dataset’s challenging annotations and variability. Notably, HED-EBT gains nearly 50\% in AP over its baseline counterpart. Similar to the above cases, the ODS and OIS also show significant gains on all models with EBT loss, approximately 3\% on average.

On UDED (Table \ref{Raw-UDED}), the average AP increases by 7.93\%. EES3-EBT achieves an AP of 0.715, outperforming its baseline by a significant 18.97\%. The average ODS and OIS are also improved by the EBT loss.

Finally, on NYUD2 (Table \ref{Raw-NYUD2}), ODS and OIS remain comparable, but AP shows a clear upward trend, with an average improvement of 19.14\%. This indicates that while edge localization remains comparable, EBT significantly enhances detection precision and stability.

Overall, the average gains across all datasets confirm the efficacy of the EBT loss. The fine-grained pixel reweighting strategy based on three distinct regions (edge, boundary, textural) enables more precise and stable edge predictions.

\begin{figure*}
    \centering
    \includegraphics[width=0.98\linewidth]{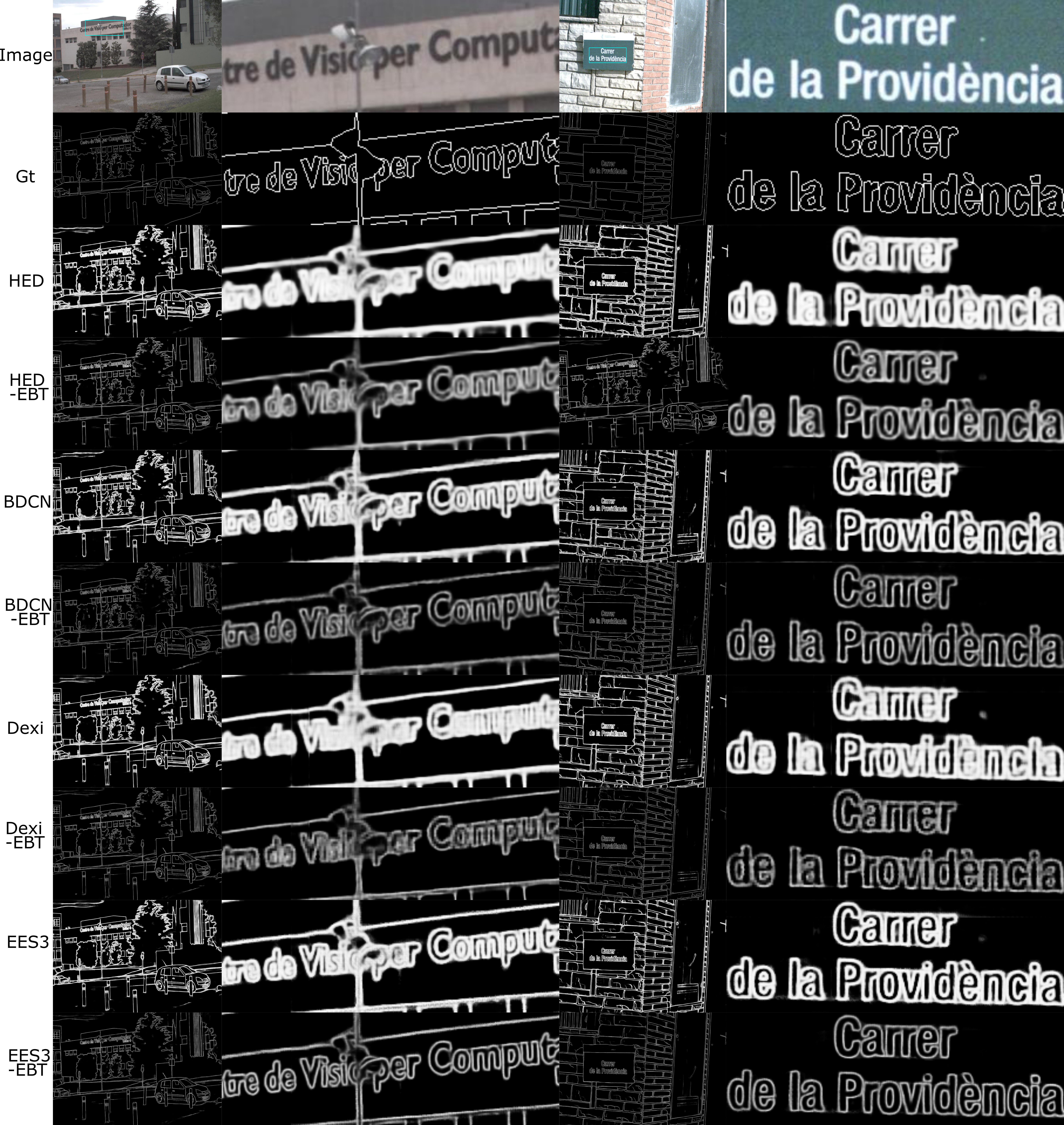}
    \caption{\ \textbf{Visual comparisons on BIPED2:} Models trained with EBT loss produce sharper edge predictions that better match ground-truth.}
    \label{BIPED2}
\end{figure*}
\begin{figure*}
    \centering
    \includegraphics[width=0.98\linewidth]{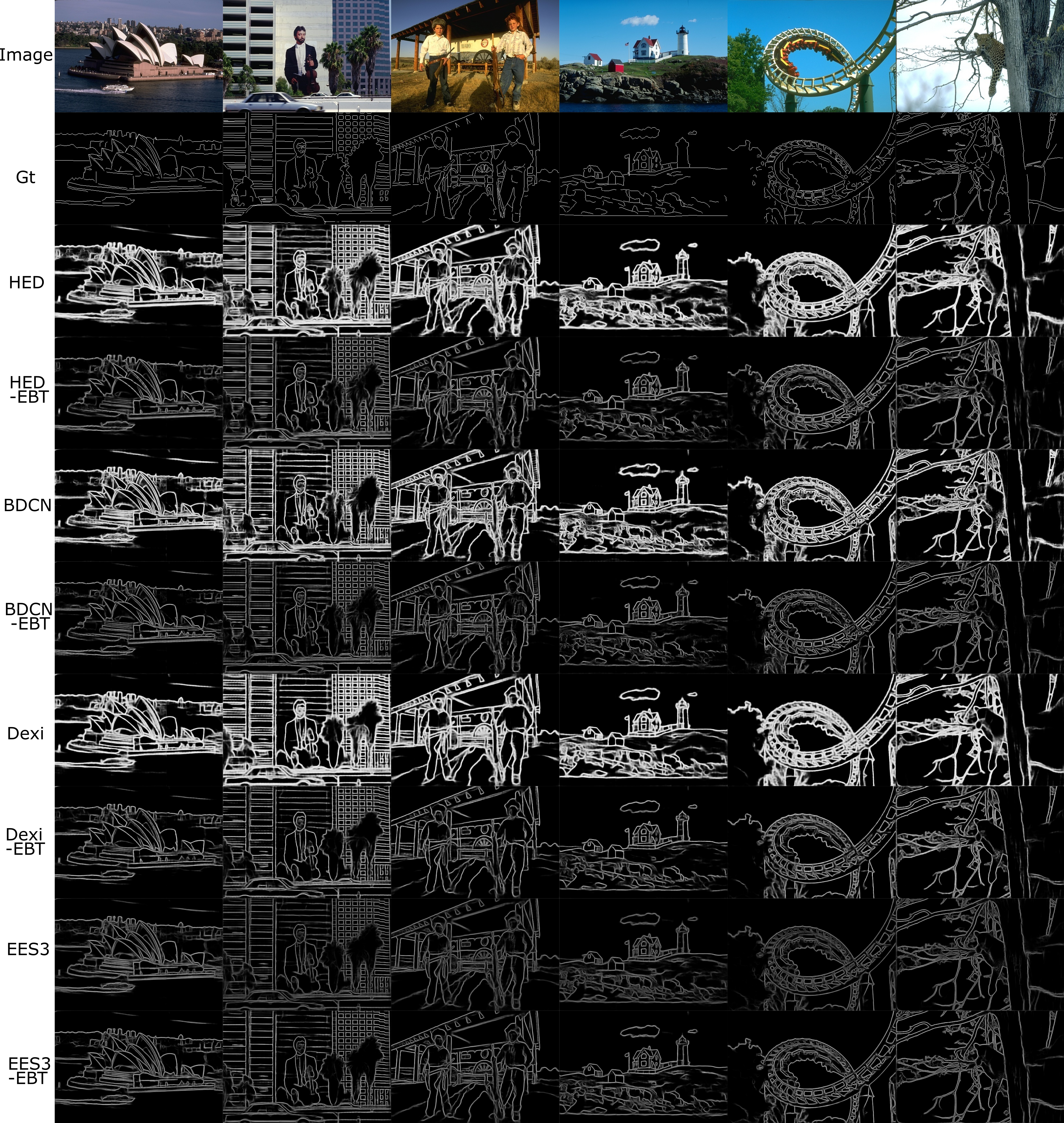}
    \caption{\ \textbf{Visual comparisons on BRIND:} Models trained with EBT loss produce sharper edge predictions that better match ground-truth.}
    \label{BRIND}
\end{figure*}

\subsubsection{Qualitative Results}

Figures~\ref{BIPED2} and~\ref{BRIND} provide visual comparisons of predictions on BIPED2 and BRIND. EBT-trained models consistently yield thinner, sharper, and more precise edge maps that align closely with the ground truth. In contrast, WBCE-trained models tend to produce slightly blurrier or thicker edges. These qualitative improvements corroborate the numerical gains observed in the AP metric.

In particular, the EBT loss enables finer distinctions in edge-neighbor areas, where standard WBCE tends to produce redundant edge responses. This validates the motivation of the EBT formulation: treating boundary pixels near edges as a separate class reduces ambiguity in labeling and leads to more perceptually meaningful edge predictions.

\subsubsection{Summary}

Experimental results demonstrate that the EBT loss generalizes effectively across models and datasets. By assigning pixel-wise weights based on edge, boundary, and texture classes, it achieves superior performance both in detection metrics and visual quality, thereby establishing it as an advanced loss function for ED. Furthermore, a unified hyperparameter set is used across all models and datasets, indicating that EBT is effective over a wide range and typically does not require extensive hyperparameter tuning.

\begin{figure*}
    \centering
    \includegraphics[width=\linewidth]{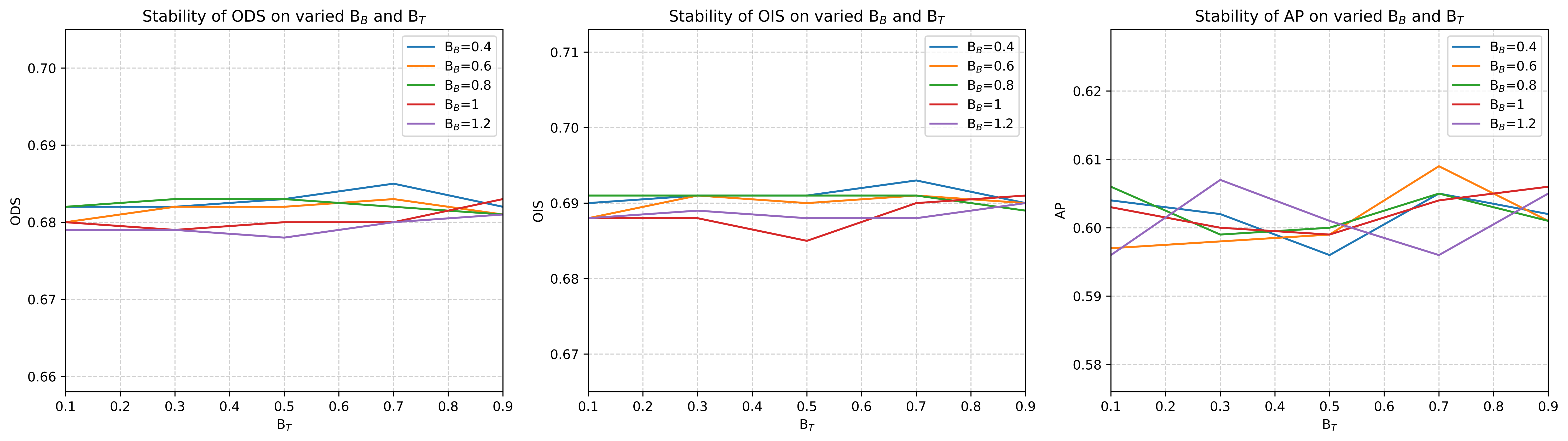}
    \caption{\ \textbf{Stability of Hyperparameter Settings:} The EES3 architecture is trained using the EBT loss under different combinations of $B_B$ and $B_T$ values, where $B_E$ is fixed at 1.0. Experiments are conducted on BRIND and evaluated using 1-pixel error tolerance without NMS. Metrics include ODS, OIS, and AP.}
    \label{Stability}
\end{figure*}

\subsection{Ablation Study}

To assess the robustness of our proposed EBT loss, we perform an ablation study focused on the stability of its hyperparameters, specifically, the balancing weights assigned to edge ($B_E$), boundary ($B_B$), and texture ($B_T$) pixels. For simplicity, we fix $B_{E}=1$ throughout the experiments, since the performance is inherently determined by the relative proportions among the three components, and normalizing one of them does not affect generality.

Figure~\ref{Stability} visualizes the performance of EES3 under varying $B_B$ and $B_T$ settings. Full numerical results are provided in the appendix. As shown, ODS, OIS, and AP remain stable across a range of parameter values, indicating that EBT is resilient to moderate hyperparameter variation and usually does not require fine-tuning to be effective.

\section{Conclusion}

In this work, we introduced the Edge-Boundary-Texture (EBT) loss, a novel training objective for ED that explicitly classifies image pixels into edge, boundary, and texture regions. Unlike the conventional WBCE loss, which formulates ED as a binary classification task, the EBT loss enforces a more structured supervision, leading to sharper and more perceptually accurate edge predictions.

We showed that the EBT loss generalizes the widely adopted WBCE loss and recovers it under extreme boundary conditions. This theoretical grounding provides a strong rationale for adopting EBT as a principled extension. By extensive experiments across diverse ED models and datasets, we demonstrated that EBT improves predictions, both quantitatively and qualitatively. Ablation studies further confirmed the robustness of the EBT loss to moderate hyperparameter changes. Notably, all main experiments used a unified set of EBT hyperparameters across models and datasets, highlighting its effectiveness without the need for fine-tuning.

These findings underscore the importance of rethinking supervision signals, particularly how non-edge pixels are treated during training, as a means to enhance the perceptual quality of ED and related tasks. Future directions include extending the EBT framework to other structured prediction tasks and exploring adaptive strategies for boundary handling.

\newpage

\section{Bibliography}
\bibliographystyle{unsrt}
\bibliography{EDBitex}

\newpage

\section{Appendix}

\subsection{Evaluations Following Traditional Benchmarks}

Tables~\ref{T-BIPED2} to \ref{T-NYUD} present the evaluation results under traditional relaxed benchmarks, where the error tolerance ranges from 4.3 to 11.1 pixels depending on image resolution. NMS is also applied in this setting.

\subsection{Data Used in the Ablation Study}

Table~\ref{Raw-Stability} provides the complete data used to generate the figures in the hyperparameter stability study described in the ablation study section. All experiments were conducted on the BRIND dataset using the EES3 architecture. The evaluation follows the same benchmark as in the main-text, namely, 1-pixel error tolerance without applying NMS.

\begin{table}[htbp]
\renewcommand\arraystretch{1.3}
\centering
\caption{\qquad \textbf{Results on BIPED2 with 11.1-pixel error tolerance with NMS:} The notations follow Table \ref{Raw-BIPED2}.}
\label{T-BIPED2}
\begin{tabular}{|p{16.25mm}<{\centering}|p{18.25mm}<{\centering}|p{18.25mm}<{\centering}|p{19.75mm}<{\centering}|}
\hline
    & ODS   & OIS   & AP \\
\hline
HED & 0.883 & 0.894 & 0.914 \\
\hline
HED-EBT & 0.883 & 0.892 & 0.919 \\
\hline
BDCN & 0.880 & 0.891 & 0.912 \\
\hline
BDCN-EBT & 0.883 & 0.890 & 0.916 \\
\hline
Dexi & 0.890 & 0.897 & 0.923 \\
\hline
Dexi-EBT & 0.885 & 0.891 & 0.904 \\
\hline
EES3 & 0.887 & 0.896 & 0.932\\
\hline
EES3-EBT & 0.890 & 0.897 & 0.929 \\
\hline
\end{tabular}
\end{table}

\begin{table}[htbp]
\renewcommand\arraystretch{1.3}
\centering
\caption{\qquad \textbf{Results on BRIND with 4.3-pixel error tolerance with NMS:} The notations follow Table \ref{Raw-BIPED2}.}
\label{T-BRIND}
\begin{tabular}{|p{16.25mm}<{\centering}|p{18.25mm}<{\centering}|p{18.25mm}<{\centering}|p{19.75mm}<{\centering}|}
\hline
    & ODS   & OIS   & AP \\
\hline
HED & 0.784 & 0.799 & 0.832 \\
\hline
HED-EBT & 0.786 & 0.801 & 0.832 \\
\hline
BDCN & 0.775 & 0.791 & 0.806 \\
\hline
BDCN-EBT & 0.793 & 0.805 & 0.839  \\
\hline
Dexi & 0.794 & 0.809 & 0.828 \\
\hline
Dexi-EBT & 0.800 & 0.812 & 0.828 \\
\hline
EES3 & 0.790 & 0.804 & 0.832 \\
\hline
EES3-EBT & 0.793 & 0.805 & 0.830 \\
\hline
\end{tabular}
\end{table}

\begin{table}[htbp]
\renewcommand\arraystretch{1.3}
\centering
\caption{\qquad \textbf{Results on BSDS500 with 4.3-pixel error tolerance with NMS:} The notations follow Table \ref{Raw-BIPED2}.}
\label{T-BSDS}
\begin{tabular}{|p{16.25mm}<{\centering}|p{18.25mm}<{\centering}|p{18.25mm}<{\centering}|p{19.75mm}<{\centering}|}
\hline
    & ODS   & OIS   & AP \\
\hline
HED & 0.622 & 0.650 & 0.578 \\
\hline
HED-EBT & 0.626 & 0.650 & 0.589 \\
\hline
BDCN & 0.628 & 0.650 & 0.596 \\
\hline
BDCN-EBT & 0.638 & 0.657 & 0.598  \\
\hline
Dexi & 0.648 & 0.672 & 0.608 \\
\hline
Dexi-EBT & 0.656 & 0.678 & 0.601 \\
\hline
EES3 & 0.650 & 0.677 & 0.643\\
\hline
EES3-EBT & 0.645 & 0.665 & 0.607 \\
\hline
\end{tabular}
\end{table}

\begin{table}[htbp]
\renewcommand\arraystretch{1.3}
\centering
\caption{\qquad \textbf{Results on UDED with 0.0075 error tolerance with NMS:} The notations follow Table \ref{Raw-BIPED2}.}
\label{T-UDED}
\begin{tabular}{|p{16.25mm}<{\centering}|p{18.25mm}<{\centering}|p{18.25mm}<{\centering}|p{19.75mm}<{\centering}|}
\hline
    & ODS   & OIS   & AP \\
\hline
HED & 0.819 & 0.857 & 0.806 \\
\hline
HED-EBT & 0.798 & 0.842 & 0.773 \\
\hline
BDCN& 0.798 & 0.837 & 0.796 \\
\hline
BDCN-EBT & 0.811 & 0.842 & 0.780 \\
\hline
Dexi& 0.829 & 0.858 & 0.824 \\
\hline
Dexi-EBT& 0.816 & 0.844 & 0.800 \\
\hline
EES3 & 0.837 & 0.862 & 0.853 \\
\hline
EES3-EBT & 0.846 & 0.870 & 0.874 \\
\hline
\end{tabular}
\end{table}

\begin{table}[htbp]
\renewcommand\arraystretch{1.3}
\centering
\caption{\qquad \textbf{Results on NYUD2 with 6-pixel error tolerance with NMS:} The notations follow Table \ref{Raw-BIPED2}.}
\label{T-NYUD}
\begin{tabular}{|p{16.25mm}<{\centering}|p{18.25mm}<{\centering}|p{18.25mm}<{\centering}|p{19.75mm}<{\centering}|}
\hline
    & ODS   & OIS   & AP \\
\hline
HED & 0.703 & 0.716 & 0.694 \\
\hline
HED-EBT & 0.696 & 0.709 & 0.648 \\
\hline
BDCN & 0.711 & 0.723 & 0.712 \\
\hline
BDCN-EBT & 0.696 & 0.709 & 0.683  \\
\hline
Dexi & 0.715 & 0.725 & 0.728 \\
\hline
Dexi-EBT & 0.717 & 0.727 & 0.687 \\
\hline
EES3 & 0.697 & 0.711 & 0.730  \\
\hline
EES3-EBT & 0.725 & 0.736 & 0.719 \\
\hline
\end{tabular}
\end{table}

\begin{table}[htbp]
\renewcommand\arraystretch{1.4}
\centering
\caption{\qquad \textbf{Full data for hyperparameter tests:} Notations follow Table \ref{Raw-BIPED2}.}
\label{Raw-Stability}
\begin{tabular}{|p{25mm}<{\centering}|p{15mm}<{\centering}|p{15mm}<{\centering}|p{15mm}<{\centering}|}
\hline
 EES3-$B_{E}$-$B_{B}$-$B_{T}$   & ODS   & OIS   & AP \\
\hline
EES3-1.0-0.4-0.1 & 0.682 & 0.690 & 0.604 \\
\hline
EES3-1.0-0.4-0.3 & 0.682 & 0.691 & 0.602 \\
\hline
EES3-1.0-0.4-0.5 & 0.683 & 0.691 & 0.596 \\
\hline
EES3-1.0-0.4-0.7 & 0.685 & 0.693 & 0.605 \\
\hline
EES3-1.0-0.4-0.9 & 0.682 & 0.690 & 0.602 \\
\hline
EES3-1.0-0.6-0.1 & 0.680 & 0.688 & 0.597 \\
\hline
EES3-1.0-0.6-0.3 & 0.682 & 0.691 & 0.598 \\
\hline
EES3-1.0-0.6-0.5 & 0.682 & 0.690 & 0.599 \\
\hline
EES3-1.0-0.6-0.7 & 0.683 & 0.691 & 0.609 \\
\hline
EES3-1.0-0.6-0.9 & 0.681 & 0.690 & 0.601 \\
\hline
EES3-1.0-0.8-0.1 & 0.682 & 0.691 & 0.606 \\
\hline
EES3-1.0-0.8-0.3 & 0.683 & 0.691 & 0.599 \\
\hline
EES3-1.0-0.8-0.5 & 0.683 & 0.691 & 0.600 \\
\hline
EES3-1.0-0.8-0.7 & 0.682 & 0.691 & 0.605 \\
\hline
EES3-1.0-0.8-0.9 & 0.681 & 0.689 & 0.601 \\
\hline
EES3-1.0-1.0-0.1 & 0.680 & 0.688 & 0.603 \\
\hline
EES3-1.0-1.0-0.3 & 0.679 & 0.688 & 0.600 \\
\hline
EES3-1.0-1.0-0.5 & 0.680 & 0.685 & 0.599 \\
\hline
EES3-1.0-1.0-0.7 & 0.680 & 0.690 & 0.604 \\
\hline
EES3-1.0-1.0-0.9 & 0.683 & 0.691 & 0.606 \\
\hline
EES3-1.0-1.2-0.1 & 0.679 & 0.688 & 0.596 \\
\hline
EES3-1.0-1.2-0.3 & 0.679 & 0.689 & 0.607 \\
\hline
EES3-1.0-1.2-0.5 & 0.678 & 0.688 & 0.601 \\
\hline
EES3-1.0-1.2-0.7 & 0.680 & 0.688 & 0.596 \\
\hline
EES3-1.0-1.2-0.9 & 0.681 & 0.690 & 0.605 \\
\hline
\end{tabular}
\end{table}

\end{document}